\documentclass[9pt]{IEEEtran}

\usepackage{caption}
\usepackage{cite}
\usepackage{titlesec}
\titlespacing*{\subsection}{0pt}{1.5ex}{1ex}
\titlespacing*{\subsubsection}{0pt}{1.5ex}{1ex}
\usepackage{amsmath,amssymb,amsfonts}
\usepackage{algorithmic}
\usepackage{graphicx}

\usepackage{textcomp}
\def\BibTeX{{\rm B\kern-.05em{\sc i\kern-.025em b}\kern-.08em
    T\kern-.1667em\lower.7ex\hbox{E}\kern-.125emX}}
\renewcommand{\thepage}{}
\begin{document}

\title{A Multi-modal Approach to Dysarthria Detection and Severity Assessment Using Speech and Text Information}

\author{\IEEEauthorblockN{1\textsuperscript{st} Anuprabha M}
\IEEEauthorblockA{\textit{Speech Processing Laboratory, LTRC} \\
\textit{IIIT Hyderabad, India}\\
anuprabha.m@research.iiit.ac.in}
\and
\IEEEauthorblockN{2\textsuperscript{nd} Krishna Gurugubelli}
\IEEEauthorblockA{\textit{Samsung Research \& Development Institute Bengaluru} \\
India\\
krishna.g@samsung.com}
\and
\IEEEauthorblockN{3\textsuperscript{rd} Kesavaraj.v}
\IEEEauthorblockA{\textit{Speech Processing Laboratory, LTRC} \\
\textit{IIIT Hyderabad, India}\\
kesavaraj.v@research.iiit.ac.in} \and

\IEEEauthorblockN{4\textsuperscript{th} Anil Kumar Vuppala}
\IEEEauthorblockA{\textit{Speech Processing Laboratory, LTRC} \\
\textit{IIIT Hyderabad, India}\\
anil.vuppala@iiit.ac.in}
}


\author{\IEEEauthorblockN{Anuprabha M\IEEEauthorrefmark{1}, Krishna Gurugubelli\IEEEauthorrefmark{2}, Kesavaraj V\IEEEauthorrefmark{1}, Anil Kumar Vuppala\IEEEauthorrefmark{1}}\\
\IEEEauthorblockA{\IEEEauthorrefmark{1}\textit{LTRC, International Institute of Information Technology-Hyderabad, India\\ \IEEEauthorrefmark{2}\textit{Samsung Research \& Development Institute-Bengaluru, India}} \\
\textit{anuprabha.m@research.iiit.ac.in, krishna.g@samsung.com, kesavaraj.v@research.iiit.ac.in, anil.vuppala@iiit.ac.in}
}}

\maketitle

\begin{abstract}
Automatic detection and severity assessment of dysarthria are crucial for delivering targeted therapeutic interventions to patients. While most existing research focuses primarily on speech modality, this study introduces a novel approach that leverages both speech and text modalities. By employing cross-attention mechanism, our method learns the acoustic and linguistic similarities between speech and text representations. This approach assesses specifically the pronunciation deviations across different severity levels, thereby enhancing the accuracy of dysarthric detection and severity assessment. All the experiments have been performed using UA-Speech dysarthric database. Improved accuracies of 99.53\% and 93.20\% in detection, and 98.12\% and 51.97\% for severity assessment have been achieved when speaker-dependent and speaker-independent, unseen and seen words settings are used. These findings suggest that by integrating text information, which provides a reference linguistic knowledge, a more robust framework has been developed for dysarthric detection and assessment, thereby potentially leading to more effective diagnoses.
\end{abstract}


\section{Introduction}

Dysarthria, a motor speech disorder that is connected with abnormalities in respiration, functioning of larynx, direction of airflow, and articulation, leads to reduced speech quality and clarity. It often occurs in various neurological disorders and is linked to progressive neurological diseases. Speakers with dysarthria, hence, face difficulties in communicating and in maintaining social connections \cite{enderby2013disorders}. Therefore, identifying and monitoring the progression of dysarthria becomes crucial for delivering appropriate therapies. Generally, dysarthric detection and severity assessment are carried out by Speech Language Pathologists (SLPs) through a series of severity assessment tests \cite{kent1989toward}. To enhance efficiency and minimize human error in these subjective assessments, it is recommended to use readily available, user-friendly objective detection and severity assessment methods \cite{gurevich2017speech}. 

In such objective assessments, it is imperative to extract meaningful features from speech. To focus on different aspects of speech such as articulation, perception, speech quality and prosody, features like MFCC, mel-spectrogram, perceptually enhanced single frequency cepstral coefficients (PE-SFCC), fundamental frequency (f0), formants, jitter, shimmer have been used \cite{chandrashekar2019spectro, gurugubelli2019perceptually, al2021classification}. Features from pre-trained Automatic Speech Recognition (ASR) models such as Wav2Vec, Hubert and Whisper have also been shown to provide improved performance as they include information about linguistic contents that pertain to speech intelligibility \cite{javanmardi2023wav2vec,rathod23_interspeech}. On the other hand, Tripathi et al.\cite{tripathi2020improved} have utilized posterior probabilities from DeepSpeech-ASR model for severity classification. Deep Neural Network based classifiers including Convolutional Neural Networks (CNN), Gated Recurrent Units (GRU) and Long Short-Term Memory networks, have demonstrated improved performance compared to traditional machine learning classifiers as they serve as both feature extractors and classifiers \cite{joshy2022automated}. 

In the investigations mentioned above, speech is the most commonly used modality for detection and assessment of dysarthria. Given the challenges involved in collecting speech data from dysarthric speakers, it is essential to leverage additional information from various modalities, if available. Apart from speech features, video-based approach \cite{bandini2016markerless}, utilizing kinematic features obtained from lip movements have also been reported to perform efficiently in detection of hypokinetic dysarthria. Another study by Xue et al. \cite{xue2023assessing} uses text based phonetic level measures such as accuracy of phoneme detected and phonetic distance, to measure the articulation patterns of dysarthic subjects. 

With advancements in speech technology, multi-modal systems have become more effective and robust at capturing complex, high-level information even when the data available is limited. Multi-modal approaches are recently being used in Speech Emotion Recognition tasks to predict emotional states by combining both audio and text modalities using cross attention \cite{khan2024mser}. Studies proposed by Tong et al. \cite{tong20b_interspeech} and Liu et al. \cite{ liu2024automatic} utilize visual features to jointly learn audio and visual information in assessing dysarthric severity.

Drawing inspiration from the aforementioned studies, this paper proposes a novel multi-modal method to leverage the words uttered by dysarthric speakers, in the form of text, in addition to speech. Incorporating text features in dysarthric assessment provides a lexical reference or a ground truth representation for standard pronunciation. This lexical reference allows the model to learn more distinguishing representations compared to traditional acoustic features, thus leading to improved performance \cite{shao22b_interspeech}. Moreover, the linguistic information embedded in the text helps the model to capture effectively, the pronunciation errors associated with difficult-to-articulate words, thereby improving accuracy in detection and severity assessment. This multi-modal approach thus enables a more comprehensive evaluation of speech intelligibility, leading to more precise diagnoses and targeted therapeutic interventions. To the best of our knowledge, this is the first time, speech and text information have been combined together for dysarthric detection and severity assessment. The main contributions of this work are summarized as follows:
\begin{itemize}
\item Proposed a novel methodology wherein text is used as a modality in addition to speech, for dysarthric detection and severity assessment.
\item Extensive systematic investigations have been conducted to compare models trained solely with speech features and with both speech and text features.
\item The models are assessed on UA-Speech dysarthric database \cite{kim2008dysarthric} in speaker-dependent (SD) and speaker-independent (SID), seen and unseen words settings.
\item This work analyses the contribution of different word groups from the UA-Speech database and highlights how text features contribute to the interpretability of the model's decision, thus providing insights into how text and speech interact in the context of dysarthria detection and severity assessment.
\end{itemize}
This paper is structured as follows: Section~II explains the proposed approach, Section~III introduces the experimental setup along with data preparation, Section~IV discusses the results and Section~V concludes the study. 

\section{Proposed approach}
In this section, we propose a novel multi-modal framework that integrates both speech (S) and text features (T) to improve accuracies in dysarthric detection and severity assessment. The core idea is to model the distribution $P(C|S,T)$ to predict the severity class (C) using a cross-attention mechanism, thereby capturing the intricate dependencies between speech and text that are crucial for identifying dysarthria. This section outlines the motivation and mathematical formulation of the proposed approach, followed by a detailed description of the architecture.

\subsection{Problem Formulation}

The proposed framework is designed to improve dysarthric detection and assessment by leveraging the complementary strengths of both speech and text modalities. Integrating text information provides a crucial reference for how the ideal utterance should be articulated \cite{shao22b_interspeech}. This enables the model to detect deviations (pronunciation errors) caused by dysarthria. The decision rule of the detection and assessment tasks is defined as follows: 
\begin{align}
C^* &=  \arg\max_{C} \{ P(C \mid S, T) \} 
\label{eq:ref_equ}
\end{align}
where $P(C|S,T)$ represents the posterior probability for estimating the severity class given both acoustic and linguistic informations. Using Bayes’ theorem,
\begin{align}
C^* &= \arg\max_{C} \left\{ \frac{P(S, T \mid C) \cdot P(C)}{P(S, T)} \right\} \\
C^* &= \arg\max_{C} \{ P(S, T \mid C) \cdot P(C) \} 
\label{eq:ref_equ}
\end{align}
where $P(S,T|C)$ represents joint-likelihood distribution and $P(C)$ represents prior probability distribution. From Equation~(3) the decision rule can be rewritten as,
\begin{align}
C^* &= \arg\max_{C} \{ P(S \mid T, C) \cdot P(T \mid C) \cdot P(C) \}
\label{eq:ref_equ}
\end{align}
The conditional distributions $P(S|T,C)$ and $P(T|C)$ can be used to find the posterior probability $P(C|S,T)$, for the detection and assessment tasks. These probabilities can capture how the acoustic realization of speech deviates from the expected norm provided the keyword phrase. Since $S$ and $T$ are conditionally dependent for each keyword, modeling the joint-likelihood $P(S,T|C)$ becomes more intricate. This can be addressed using multivariate Gaussian distributions or multi-modal neural networks. In the proposed multi-modal framework, inputs $S$ and $T$ are processed by distinct encoders and their outputs are jointly processed using a cross-attention mechanism. This mechanism explicitly captures the dependencies between the two modalities before computing the class logits, enabling the model to leverage the interactions between $S$ and $T$ for improved classification.




\begin{figure}[h]
      \centering
      \includegraphics[width=0.7\linewidth]{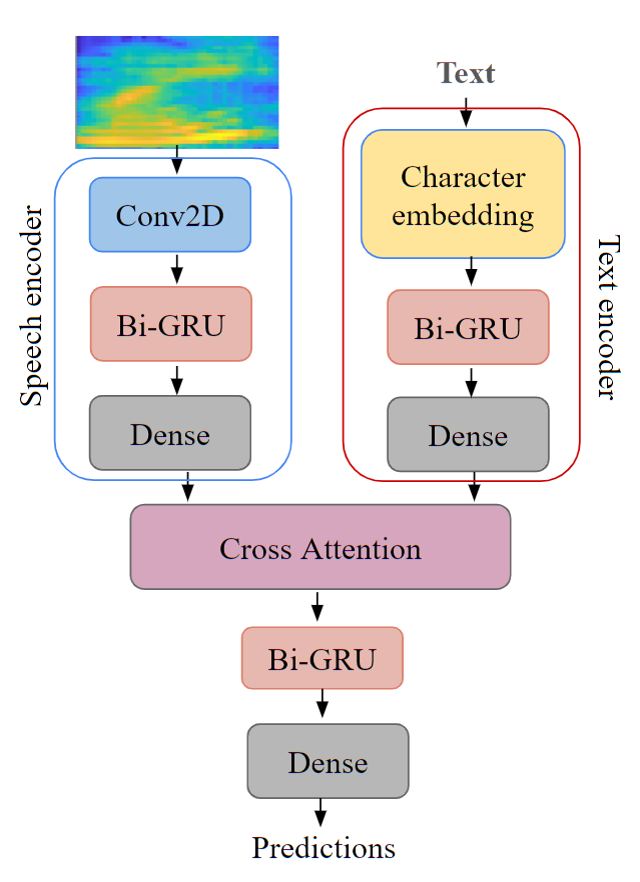}
      \caption { Proposed architecture for detection and severity assessment of dysarthria}
\end{figure}

\subsection{Proposed Architecture}
To model the likelihood distribution $P(S,T|C)$, we propose an architecture as shown in Fig.~1, that consists of three main components: a speech encoder, a text encoder, and a cross-attention layer. Together, these components learn a shared representation that captures the similarity between the acoustic features of speech and the linguistic representations of text.

\subsubsection{Speech Encoder} The speech encoder receives time-frequency representation of speech in the form of mel-spectrogram which essentially captures the spectral patterns associated with the speech characteristics \cite{chandrashekar2019spectro}. The CNN layer and bi-directional GRU (Bi-GRU) are employed to capture low-level features and overall temporal variations present in the speech signal. In this way, the speech embeddings obtained from the speech encoder provide information about the subtle variations in how the word is articulated as well as the overall challenges in uttering the word. These speech embeddings are denoted by the conditional distribution $P(S|T,C)$ which allows the speech encoder to learn diverse speech patterns irrespective of the severity. 

\subsubsection{Text Encoder} The text encoder receives character level sequences for any given word. To obtain character level tokenization for processing isolated words, an embedding layer with a vocabulary size of 26 (corresponding to the number of distinct characters in English) is employed. These character level embeddings are processed by Bi-GRU to obtain a fixed-dimensional representation, regardless of the number of characters present in any keyword. Text embeddings, thus obtained, provide a unique representation for the character sequence present in each keyword. These text embeddings act as a linguistic representation of speech for each keyword against which speech utterances belonging to different severity levels are compared. 

\subsubsection{Multi-modal Classifier} Cross-attention mechanism \cite{vaswani2017attention} is employed to obtain a shared representation between speech and text embeddings, and is modeled by the likelihood function $P(S,T|C)$. The final representation thus obtained retains the cues relevant to speech severity ranging from healthy to high level of dysarthric severity, depending on the task. Finally, necessary dense layers and activation functions are employed to carryout detection and severity assessment tasks.

\section{Experimental setup}
This section describes the database used, details about the different settings and the implementation details related to the model training. 

\subsection{Database}
The experiments in this study are conducted using UA-Speech dysarthric database in English  \cite{kim2008dysarthric}. This database contains speech samples from 15 dysarthric speakers with cerebral palsy (CP) and 11 healthy speakers. The database contains isolated utterances of digits (10), radio alphabets (26), computer commands (19), common words~(100) from the Brown corpus and uncommon words (300) selected from Project Gutenberg. There are 3 blocks in the database, each block has 100 uncommon words and 155 words from other categories mentioned above. Based on the speech intelligibility ratings provided by native listeners, the severity is categorized into 4 groups; very low, low, medium and high. 
The speech samples recorded from array~6 sampled at 16 kHz are used for the analysis. In the database, duration of the speech samples of dysarthric speakers are varied from less than 2 seconds to greater than 18 seconds, depending on the severity of the speakers. For this experiment, samples with duration up to 10 seconds are taken after trimming silences from the leading and trailing ends. For this study, mel-spectrograms generated by extracting 80-dimensional mel-filter bank coefficients from speech for every 10~ms using a 25~ms window, are used as speech features with dimensions (N × 80), where N is the number of frames in each speech sample.

\subsection{Comprehensive Details of Different Settings} 
\label{sec:pagestyle}
To study the effect of incorporating text knowledge, both detection and severity classification tasks are carried out using speaker-dependent and speaker-independent settings, as well as seen and unseen words settings, namely, SD, SID-1 and SID-2. To include different articulation patterns \cite{kim2008dysarthric} in training the models, randomly sampled unseen words are included in the train set. For each of these settings, the distribution of common and uncommon words in the training and test sets are detailed in TABLE~I. For detection task, 11 healthy speakers and 15 dysarthric speakers have been considered. 26 models are trained for each SID setting by following Leave One Speaker Out (LOSO) cross validation approach. For severity assessment task, as the speaker distribution is uneven across the classes, 8 speakers are considered for training and remaining 7 speakers are used for testing. This methodology has been adapted from \cite{joshy2023dysarthria} to ensure equal number of speakers across severity levels during training. Speech-only detection and assessment models are also employed to compare with the performance of the proposed multi-modality detection and assessment models.

\begin{table}[h]
\centering
\caption{Distribution of common and uncommon words used in train and test sets of dysarthric detection and severity assessment tasks for different settings: speaker-dependent and unseen words (SD), speaker-independent seen words (SID-1), and unseen words (SID-2).}
\label{tab:my-table}
\renewcommand{\arraystretch}{1.2}
\resizebox{\columnwidth}{!}{%
\begin{tabular}{|c|c|c|}
\hline
Experimental setting & Train set & Test set  \\ \hline
\hline
SD    & \begin{tabular}[c]{@{}c@{}}155 common words and \\ 200 uncommon words\end{tabular} & 100 uncommon words \\ \hline
SID-1                & All words & All words \\ \hline
SID-2 & \begin{tabular}[c]{@{}c@{}}155 common words and \\ 200 uncommon words\end{tabular} & 100 uncommon words \\ \hline
\end{tabular}%
}
\end{table}

\subsection{Implementation Details}
\label{sec:pagestyle}


The speech encoder shown in Fig. 1 consists of two 2D CNN layers to process 2 dimensional mel-spectrograms. To prevent over-fitting and to ensure stable training, a dropout layer of 0.2 and batch normalization are applied after each CNN layer. Two Bi-GRUs (dimension of 64) followed by dense layers are applied in both speech and text encoders to obtain fixed dimensional speech and text features, which are then passed on to the cross-attention module. In cross-attention setup, speech embedding acts as both the key and value, while the text embedding acts as the query. The context vectors obtained from cross-attention are processed by GRU (dimension of 64), and dense layers (128 and 32). Finally, a sigmoid activation function is applied to detect dysarthric speech, while a softmax activation function is employed to assess severity levels (very low, low, medium and high) present in the UA-Speech database. To obtain predictions for speech-only model, speech encoder alone is considered. Conversely, for multi-modality model, the predictions are evaluated after applying cross-attention.

Binary and categorical cross-entropy loss are utilised as training criteria for binary and multi-class classification, respectively, in conjunction with Adam optimizer with initial learning rate of $10^{-4}$. To dynamically adjust the learning rate, ReduceOnPlateau strategy is applied with patience parameter of 5 epochs. In addition, an early stopping mechanism with a patience of 3 epochs is also applied to prevent over-fitting while training. For training, a NVIDIA Ge-Force RTX 2080 Ti GPU has been used.

\section{Results and Discussion}
\label{sssec:subsubhead}
This study investigates the robustness of the proposed multi-modality approach for dysarthric detection and severity assessment. Extensive experimental analyses have been conducted across various settings, as detailed in TABLE~I. The experiments utilized the UA-Speech dataset, described in Section~III, and the results are presented in terms of accuracy.


\subsection{Comparative Analysis of Speech-Only and Speech-Text Modalities} 
This study compares speech-only and speech-text modalities in performance assessment of two tasks: dysarthric speech detection and severity assessment, across various experimental settings (as detailed in TABLE~I). The results, presented in TABLE~II, demonstrate that in SD setting, incorporating text features yields an absolute improvement of 4.15\% and 0.62\% in accuracy, for detection and assessment tasks, respectively, compared to the speech-only modality. In the SID-1 setting which is preferred by clinicians for diagnosing new patients, the speech-text modality shows an absolute improvement of 3.59\% and 2.83\% in accuracy, for detection and assessment tasks, respectively. This clearly indicates the model's ability to learn dysarthric related information even when it is tested on unknown speakers using seen words. Conversely, when the model is tested using unseen words, in the SID-2 setting, a performance decline of 2.46\%  in detection accuracy is observed, in the speech-text modality. Interestingly, for severity classification using the same SID-2 setting, an improvement of 2.71\% in accuracy is observed which suggests enhanced generalization to both unknown speakers and unseen words.


\begin{table}[h]
\centering
\footnotesize
\caption{Performance (accuracy in \%) comparison of the proposed multi-modality (speech-text) model with speech-only model for dysarthric detection and assessment tasks across different settings: speaker-dependent and unseen words (SD), speaker-independent seen words (SID-1), and unseen words (SID-2).}
\label{tab:my-table}
\renewcommand{\arraystretch}{1.5}
\resizebox{\columnwidth}{!}{%

\begin{tabular}{|c|ccc||ccc|}
\hline

         & \multicolumn{3}{c||}{Dysarthric detection}                                  & \multicolumn{3}{c|}{Severity assessment}                            \\ \hline
\hline
Modality & \multicolumn{1}{c|}{SD}    & \multicolumn{1}{c|}{SID-1} & SID-2        & \multicolumn{1}{c|}{SD}    & \multicolumn{1}{c|}{SID-1} & SID-2 \\ \hline
Speech-only   & \multicolumn{1}{c|}{95.38} & \multicolumn{1}{c|}{89.61}   & \textbf{87.76} & \multicolumn{1}{c|}{97.50} & \multicolumn{1}{c|}{49.14}   & 54.71   \\ \hline
Speech-text &
  \multicolumn{1}{c|}{\textbf{99.53}} &
  \multicolumn{1}{c|}{\textbf{93.20}} &
  85.30 &
  \multicolumn{1}{c|}{\textbf{98.12}} &
  \multicolumn{1}{c|}{\textbf{51.97}} &
  \textbf{57.42} \\ \hline
\end{tabular}
}
\end{table}

\begin{table*}[t!]
\centering
\caption{Performance (accuracy in \%) for different word groups under speaker-independent (SID-1) setting for both speech-only and speech-text modalities in dysarthric detection and severity assessment tasks. }

\label{tab:my-table}
\scriptsize
\renewcommand{\arraystretch}{1.5}

\begin{tabular}{|ccccccccccccc|}
\hline
\multicolumn{1}{|c|}{Word groups} &
  \multicolumn{1}{c|}{Digits} &
  \multicolumn{1}{c|}{Commands} &
  \multicolumn{1}{c|}{\begin{tabular}[c]{@{}c@{}}Alphabets\end{tabular}} &
  \multicolumn{1}{c|}{\begin{tabular}[c]{@{}c@{}}Common \end{tabular}} &
  \multicolumn{1}{c|}{\begin{tabular}[c]{@{}c@{}}Uncommon \end{tabular}} &
  \multicolumn{1}{c|}{\begin{tabular}[c]{@{}c@{}}B1\\ uncommon\end{tabular}} &
  \multicolumn{1}{c|}{\begin{tabular}[c]{@{}c@{}}B2\\ uncommon\end{tabular}} &
  \multicolumn{1}{c|}{\begin{tabular}[c]{@{}c@{}}B3\\ uncommon\end{tabular}} &
  \multicolumn{1}{c|}{B1\_all} &
  \multicolumn{1}{c|}{B2\_all} &
  \multicolumn{1}{c|}{B3\_all} &
  All words \\ \hline
  \hline
\multicolumn{13}{|c|}{\textbf{Dysarthric speech detection (Healthy Vs Dysarthria)}} \\ \hline
\hline
\multicolumn{1}{|c|}{\begin{tabular}[c]{@{}c@{}}Speech-only \\ \end{tabular}} &
  \multicolumn{1}{c|}{91.28} &
  \multicolumn{1}{c|}{90.80} &
  \multicolumn{1}{c|}{91.51} &
  \multicolumn{1}{c|}{91.12} &
  \multicolumn{1}{c|}{87.20} &
  \multicolumn{1}{c|}{87.16} &
  \multicolumn{1}{c|}{87.83} &
  \multicolumn{1}{c|}{86.65} &
  \multicolumn{1}{c|}{90.10} &
  \multicolumn{1}{c|}{89.64} &
  \multicolumn{1}{c|}{89.10} &
  89.61 \\ \hline
\multicolumn{1}{|c|}{\begin{tabular}[c]{@{}c@{}}Speech-text\\  \end{tabular}} &
  \multicolumn{1}{c|}{95.25} &
  \multicolumn{1}{c|}{94.55} &
  \multicolumn{1}{c|}{95.26} &
  \multicolumn{1}{c|}{95.03} &
  \multicolumn{1}{c|}{90.36} &
  \multicolumn{1}{c|}{90.58} &
  \multicolumn{1}{c|}{90.82} &
  \multicolumn{1}{c|}{92.64} &
  \multicolumn{1}{c|}{93.61} &
  \multicolumn{1}{c|}{93.37} &
  \multicolumn{1}{c|}{92.64} &
  93.20 \\ \hline
  \hline
\multicolumn{13}{|c|}{\textbf{Dysarthric severity assessment (very low Vs low Vs medium Vs high)}} \\ 
\hline

\multicolumn{1}{|c|}{\begin{tabular}[c]{@{}c@{}}Speech-only \\ \end{tabular}} &
  \multicolumn{1}{c|}{44.23} &
  \multicolumn{1}{c|}{48.73} &
  \multicolumn{1}{c|}{44.44} &
  \multicolumn{1}{c|}{46.91} &
  \multicolumn{1}{c|}{53.23} &
  \multicolumn{1}{c|}{53.73} &
  \multicolumn{1}{c|}{53.61} &
  \multicolumn{1}{c|}{52.34} &
  \multicolumn{1}{c|}{50.28} &
  \multicolumn{1}{c|}{47.20} &
  \multicolumn{1}{c|}{49.97} &
  49.14 \\ \hline
\multicolumn{1}{|c|}{\begin{tabular}[c]{@{}c@{}}Speech-text \\ \end{tabular}} &
  \multicolumn{1}{c|}{46.71} &
  \multicolumn{1}{c|}{53.84} &
  \multicolumn{1}{c|}{45.14} &
  \multicolumn{1}{c|}{47.23} &
  \multicolumn{1}{c|}{56.25} &
  \multicolumn{1}{c|}{56.02} &
  \multicolumn{1}{c|}{58.06} &
  \multicolumn{1}{c|}{54.63} &
  \multicolumn{1}{c|}{51.09} &
  \multicolumn{1}{c|}{50.16} &
  \multicolumn{1}{c|}{50.61} &
  51.47 \\ \hline
\end{tabular}%
\end{table*}

Based on the above findings it is evident that the multi-modality approach surpasses speech-only models, as the additional text features which provide valuable linguistic information serve as a reference for how the speech should be. Since the model is provided with these text references it learns deviations in the speech caused by slurring or mispronunciation in dysarthric speakers.

\subsection{Performance Assessment of Proposed Framework Across Diverse Word Groups} 
The performance of the proposed framework may be influenced by the specific input text information. Therefore, we conducted a detailed analysis of the framework across different word groups and the findings are presented in TABLE~III. The results highlight that while the performance trends of both the speech-only and speech-text models are consistent across different word groups, the multi-modality approach consistently outperforms the speech-only model in terms of accuracy due to incorporation of text features as discussed earlier. For instance, considering common and uncommon word groups, the absolute improvement in accuracies for the multi-modality model compared to speech-only model are 3.91\% and 3.16\% for detection, 0.32\% and 3.02\% for severity assessment, respectively. 

For a given modality, across the various word groups indicated in TABLE~III, a marginal variation in performance has been observed. This can be attributed to the specific characteristics of dysarthria as well as the phonetic properties of the individual word groups themselves. Detection accuracy is notably higher for common words, digits and computer commands which are likely to be easier for dysarthric speakers to articulate due to their familiarity and reduced articulatory demands. Even words that need less articulatory movements provide sufficient information for the model to effectively distinguish between healthy and dysarthric speakers. This is clearly shown in terms of the absolute improvement of 4.67\% in accuracy for common word group compared to uncommon word group, for the multi-modality model.

Unlike detection, for severity classification, the multi-modality model performs better for uncommon words such as ‘advantageous’ and ‘boulevard’ in the UA-Speech database. These words require more complex articulatory movements \cite{tripathi2020novel} and are challenging for dysarthric speakers to pronounce due to potential pauses, repeating phonemes, and other complexities. In such a scenario, the model exhibits an absolute improvement of 9.02\%  in accuracy compared to the common word group. This can be attributed to the fact that the model leverages the linguistic information embedded in the text to better capture and analyze the pronunciation errors associated with these difficult-to-articulate words, thereby improving severity classification accuracy.

\subsection{Comparison with State-of-the-Art Systems}
The performance of the proposed system is benchmarked against various state-of-the-art systems, with results presented in TABLE~IV. To ensure a fair comparison, we selected models evaluated under similar training and testing conditions. In severity assessment, when compared with SECNN model trained with spectrograms \cite{joshy2023dysarthria}, the proposed multi-modality model shows an absolute improvement of 0.54\% and 0.19\% in terms of accuracy for both SD and SID. Also in detection, compared to model trained with Wav2Vec features \cite{javanmardi2023wav2vec}, proposed model shows an absolute improvement of 1.72\% for SID setting. The comparison clearly demonstrates that the proposed speech-text multi-modality models outperform the speech-only models. This improvement is attributed to the additional linguistic context provided by the text features, which enhances the model's ability to detect dysarthria and assess its severity by analyzing deviations in pronunciation, effectively.

\begin{table}[]
\centering
\caption{Performance (accuracy in \%) comparison of the proposed multi-modality approach with speech-only state-of-the-art approaches. }
\setlength{\tabcolsep}{4.5 pt}
\renewcommand{\arraystretch}{1.3}
\label{tab:my-table}
\resizebox{\columnwidth}{!}{%
\begin{tabular}{|cccc|}
\hline
\multicolumn{1}{|c|}{Features and models}     & \multicolumn{1}{c|}{SD}    & \multicolumn{1}{c|}{SID-1} & SID-2 \\ \hline
\hline
\multicolumn{4}{|c|}{\textbf{Dysarthric detection}}                                                                      \\ \hline
\hline
\multicolumn{1}{|c|}{PE-SFCC + i-vector with PLDA scoring \cite{gurugubelli2019perceptually}} &
  \multicolumn{1}{c|}{-} &
  \multicolumn{1}{c|}{-} &
  \textbf{93.64} \\ \hline
\multicolumn{1}{|c|}{VMD + CNN \cite{rajeswari2022dysarthric}}               & \multicolumn{1}{c|}{95.95} & \multicolumn{1}{c|}{-}     & -     \\ \hline
\multicolumn{1}{|c|}{Glottal flow +CNN+MLP \cite{narendra2020glottal}}   & \multicolumn{1}{c|}{-}     & \multicolumn{1}{c|}{87.93} & -     \\ \hline
\multicolumn{1}{|c|}{wav2vec features +  SVM \cite{javanmardi2023wav2vec}}   & \multicolumn{1}{c|}{-}     & \multicolumn{1}{c|}{93.95} & -     \\ \hline
\multicolumn{1}{|c|}{\begin{tabular}[c]{@{}c@{}}Proposed Speech-Text modality\end{tabular}} &
  \multicolumn{1}{c|}{\textbf{99.53}} &
  \multicolumn{1}{c|}{\textbf{93.20}} &
  85.30 \\ \hline
  \hline
\multicolumn{4}{|c|}{\textbf{Dysarthric severity assessment }}                                       \\ \hline
\hline
\multicolumn{1}{|c|}{Melspectrogram + TF-CNN \cite{chandrashekar2019spectro}} & \multicolumn{1}{c|}{90.25} & \multicolumn{1}{c|}{-}     & 56.50 \\ \hline

\multicolumn{1}{|c|}{Deep speech posterior + SVM \cite{tripathi2020improved}} & \multicolumn{1}{c|}{97.4} & \multicolumn{1}{c|}{-}     &  53.9  \\ \hline

\multicolumn{1}{|c|}{Melspectrogram + SECNN \cite{joshy2023dysarthria}}  & \multicolumn{1}{c|}{97.58} & \multicolumn{1}{c|}{-}     & 57.23 \\ \hline

\multicolumn{1}{|c|}{HuBERT features + CNN \cite{javanmardi2024pre}  } &   \multicolumn{1}{c|}{-} &  \multicolumn{1}{c|}{ 48.01} &  - \\ \hline

\multicolumn{1}{|c|}{\begin{tabular}[c]{@{}c@{}}Proposed Speech-Text modality\end{tabular}} &
  \multicolumn{1}{c|}{\textbf{98.12}} &
  \multicolumn{1}{c|}{\textbf{51.97}} &
  \textbf{57.42} \\ \hline
\end{tabular}%
}
\end{table}
\label{sec:typestyle}

\section{Conclusion}
{\centering\scshape}
\label{sssec:subsubhead}
This study demonstrates that the proposed multi-modality model, which integrates both speech and text, provides a significant advantage over speech-only model in detecting and assessing dysarthria. By leveraging the complementary strengths of both acoustic and linguistic knowledge, this multi-modality model achieves greater accuracy and consistency, particularly when dealing with the complex phonetic properties associated with dysarthric speech. Notably, in speaker-independent settings which are more preferred in clinical environment, this model shows higher accuracies of 93.20\% and 51.97\% in dysarthric detection and severity assessment, respectively. Additionally, this study provides insights into the influence of articulatory patterns having different difficulty levels, on the performance of the model. 
In the future work, the focus will be on employing different strategies to combine various speech and text representations towards developing more robust models for diagnosis of dysarthria.

\newpage

\bibliographystyle{IEEEbib}
\bibliography{refs}

\end{document}